\documentclass[manuscript,screen,review=false]{acmart}

\AtBeginDocument{%
  \providecommand\BibTeX{{%
    \normalfont B\kern-0.5em{\scshape i\kern-0.25em b}\kern-0.8em\TeX}}}

\usepackage{subfig}

\usepackage{wrapfig}
\newcommand{\Lagr}{\mathcal{L}}

\usepackage{caption}

\usepackage{algorithm,algpseudocode}
\usepackage{microtype}

\usepackage{ragged2e}
\usepackage{multirow}
\usepackage{graphicx}

\acmConference[Conference acronym 'XX]{Make sure to enter the correct
  conference title from your rights confirmation emai}{June 03--05,
  2018}{Woodstock, NY}
\acmPrice{15.00}
\acmISBN{978-1-4503-XXXX-X/18/06}

\begin{document}

\title{\textit{XploreNAS}: Explore Adversarially Robust \& Hardware-efficient Neural Architectures for Non-ideal Xbars}

\author{Abhiroop Bhattacharjee}
\affiliation{%
  \institution{Yale University}
  \city{New Haven}
  \country{USA}
  }
\email{abhiroop.bhattacharjee@yale.edu}

\author{Abhishek Moitra}
\affiliation{%
  \institution{Yale University}
  \city{New Haven}
  \country{USA}
  }
\email{abhishek.moitra@yale.edu}

\author{Priyadarshini Panda}
\affiliation{%
  \institution{Yale University}
  \city{New Haven}
  \country{USA}
  }
\email{priya.panda@yale.edu}

\renewcommand{\shortauthors}{Bhattacharjee, et al.}

\begin{abstract}
 Compute In-Memory platforms such as memristive crossbars are gaining focus as they facilitate acceleration of Deep Neural Networks (DNNs) with high area and compute-efficiencies. However, the intrinsic non-idealities associated with the analog nature of computing in crossbars limits the performance of the deployed DNNs. Furthermore, DNNs are shown to be vulnerable to adversarial attacks leading to severe security threats in their large-scale deployment. Thus, finding adversarially robust DNN architectures for non-ideal crossbars is critical to the safe and secure deployment of DNNs on the edge. This work proposes a two-phase algorithm-hardware co-optimization approach called \textit{XploreNAS} that searches for hardware-efficient \& adversarially robust neural architectures for non-ideal crossbar platforms. We use the one-shot Neural Architecture Search (NAS) approach to train a large Supernet with crossbar-awareness and sample adversarially robust Subnets therefrom, maintaining competitive hardware-efficiency. Our experiments on crossbars with benchmark datasets (SVHN, CIFAR10 \& CIFAR100) show upto $\sim8-16\%$ improvement in the adversarial robustness of the searched Subnets against a baseline ResNet-18 model subjected to crossbar-aware adversarial training. We benchmark our robust Subnets for Energy-Delay-Area-Products (EDAPs) using the Neurosim tool and find that with additional hardware-efficiency driven optimizations, the Subnets attain $\sim1.5-1.6\times$ lower EDAPs than ResNet-18 baseline.
\end{abstract}

\ccsdesc[500]{Hardware~Emerging architectures}
\ccsdesc[300]{Hardware~Safety critical systems}

\keywords{Neural Architecture Search, Memristive Crossbars, Non-idealities, Adversarial Robustness, EDAP}

\maketitle

\vspace{-3mm}
\section{Introduction}

\vspace{-3mm}
\label{sec:intro}

\begin{wrapfigure}{l}{0.35\textwidth}
\includegraphics[width=0.35\textwidth]{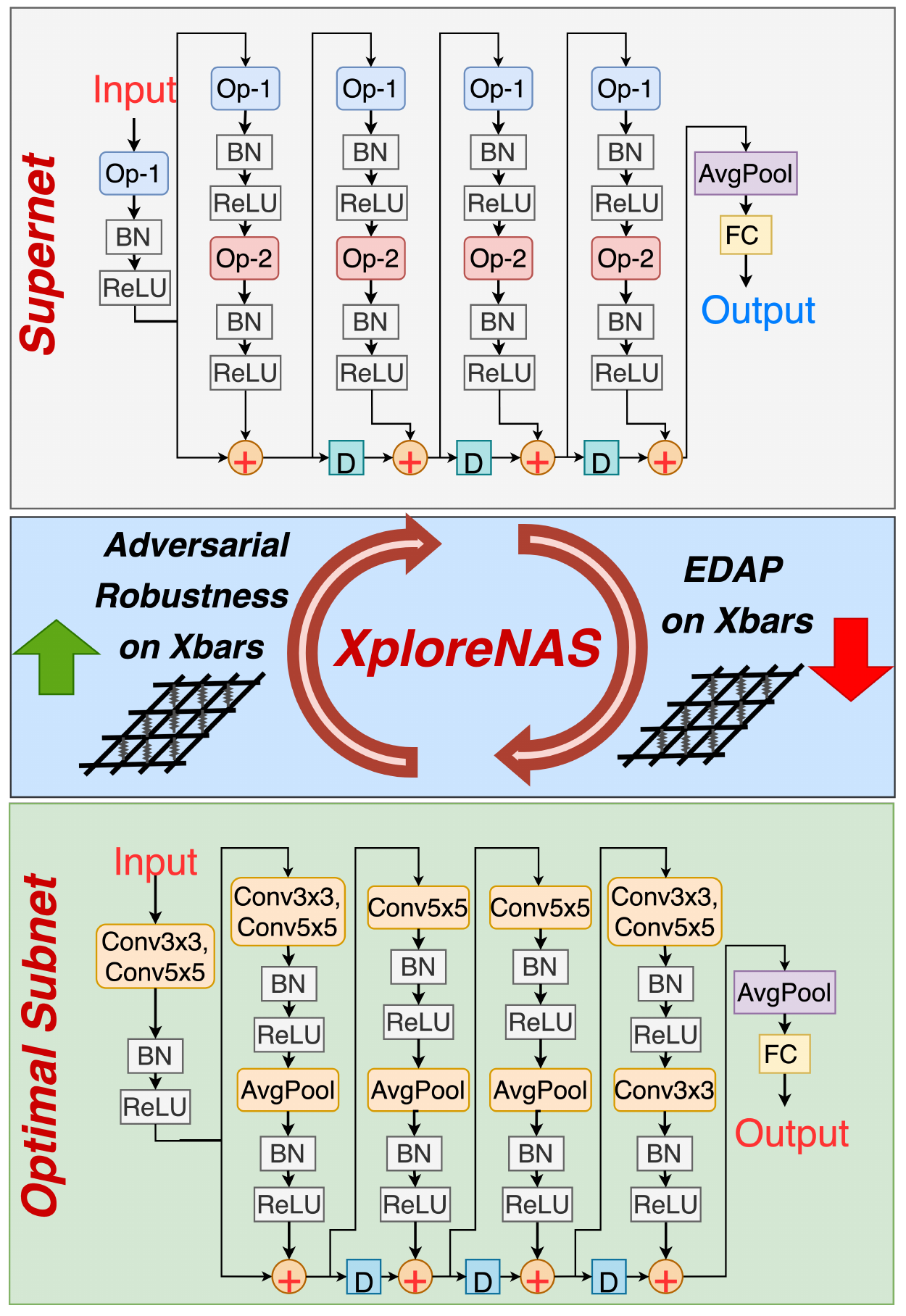}
  
\caption{Pictorial depiction of the flow of \textit{XploreNAS} that performs efficiency-driven crossbar-aware NAS to ameliorate the adversarial accuracy of DNNs on hardware.}
\label{xplorenas}
 \vspace{-8mm}
\end{wrapfigure}

Today, Neural Architecture Search (NAS) has achieved great feats in reducing human effort for designing neural architectures specific to a plethora of tasks, such as, image classification, image segmentation, natural language processing among others \cite{cai2018efficient, zoph2018learning, nekrasov2020architecture}. In the recent years, there has been a growing interest in using NAS as a tool to explore high performance and hardware-efficient deep neural networks (DNNs) tailored for platforms such as CPUs, GPUs, FPGAs and Compute In-Memory (CIM) architectures \cite{cai2018proxylessnas, cai2019once, jiang2020hardware, yang2020co, negi2021nax}. The goal is to maximize accuracy of the NAS-derived DNN models on these hardware platforms while achieving increased energy-efficiency, throughput and area-utilization. 

Memristive crossbar-based CIM has become popular as an alternative platform addressing the `memory-wall' bottleneck of von-Neumann architectures. Although memristive crossbars have the ability to perform compact and energy-efficient Multiply-and-Accumulate (MAC) operations with high throughput, they are susceptible to various non-idealities owing to their analog nature of computing \cite{hu2016dot, hu2018memristor, jain2020rxnn, chakraborty2020geniex, bhattacharjee2022examining, bhattacharjee2022examining1}. These non-idealities include interconnect parasitics, memristor device variations, etc. that degrade the performance (accuracy) of the DNN architecture mapped onto crossbars. In fact, the adversarial security of DNNs is highly compromised on CIM platforms that are prone to non-idealities \cite{bhattacharjee2020switchx, bhattacharjee2021neat, roy2020robustness, moitra2021detectx}. Adversarial attacks are structured, yet, small perturbations on the input, that fool a DNN causing high confidence misclassification \cite{goodfellow2014generative, goodfellow2014explaining}.  Today, addressing adversarial vulnerabilities of DNNs has become an active area for research \cite{goodfellow2014explaining, madry2017towards}. There have been recent works that suggest mitigation of non-idealities to improve the adversarial robustness of crossbar-mapped DNN models \cite{bhattacharjee2021neat, roy2020robustness}. But, these strategies are DNN architecture and underlying crossbar platform-specific requiring expensive re-training of the model when the crossbar size or related specifications change. No prior work has focused on obtaining DNN architectures that are adversarially robust in presence of non-idealities for multiple crossbar sizes, while maintaining competitive hardware-efficiency.

In this work, we propose \textit{XploreNAS}, a crossbar-aware NAS method that searches for optimal hardware-efficient and adversarially robust neural architectures (Subnets) for CIM platforms from a parent Supernet. We propose a two-phase hardware-aware Supernet training methodology that identifies a series of operations or paths within the Supernet architecture that collectively provide resilience against the impact of intrinsic crossbar-level noise and white-box adversarial perturbations.  
In addition to unleashing non-ideality aware adversarial robustness, \textit{XploreNAS} allows efficiency-driven hardware optimizations during the architecture search phase to find Subnets that are resource-efficient on crossbar-based platforms and expend low \textit{Energy-Delay-Area-Product} (EDAP) on hardware (see Fig. \ref{xplorenas}). 

In summary, the key contributions of our work are as follows:

\begin{itemize}
    
     \item We propose a two-phase NAS approach based on algorithm-hardware co-optimization, called \textit{XploreNAS}, to build adversarially robust DNN architectures suited for non-ideal memristive crossbar arrays.
    
    \item We conduct our experiments on non-ideal crossbars with sizes ranging from 32$\times$32 to 128$\times$128 and show that our Subnets are upto $\sim8-16\%$ more robust than a baseline adversarially trained ResNet-18 model against strong adversarial attacks.
    
    \item We introduce hardware-efficiency driven optimizations during \textit{XploreNAS} to implement highly efficient Subnets on crossbars achieving $\sim1.5-1.6\times$ lower EDAPs (benchmarked using the Neurosim tool \cite{chen2018neurosim}) than a baseline ResNet-18 architecture, in addition to being adversarially robust. 
    
    \item We also find that \textit{XploreNAS}-derived Subnets can show significant adversarial resilience over a range of crossbar sizes and device variations during inference without the need for re-training. 
    
\end{itemize}

The remainder of the work is organized as follows. Section \ref{sec:related_work} discusses related works in the areas of NAS-based solutions to relieve the adversarial accuracy of DNNs on software as well as non-NAS works based on non-ideality mitigation in CIM platforms to better the robustness of crossbar-mapped DNNs. In Section \ref{sec:background}, we present some background about adversarial attack generation, followed by depiction of dot-product computations in non-ideal memristive crossbars. Thereafter, we provide a detailed description of the \textit{XploreNAS} methodology, referred to int Fig. \ref{xplorenas}, followed by Subnet selection and fine-tuning in Section \ref{sec:method}. Section \ref{sec:metrics} enlists all the evaluation metrics to assess the performance of our \textit{XploreNAS}-derived Subnets against corresponding baselines. Section \ref{sec:expt} presents our experimental setup followed by results and discussion. In Section \ref{sec:comparison}, we present the contributions of \textit{XploreNAS} in regard to related works in a tabular form and finally, conclude our work in Section \ref{sec:conclusion}.

\section{Related Works}
\label{sec:related_work}

Works such as Guo et al. \cite{guo2020meets} have proposed a fully software-based solution to train an over-parameterized network (called Supernet) using one-shot NAS \cite{liu2017learning} and sample multiple Subnets therefrom. These Subnets are fine-tuned to yield state-of-the-art adversarial robustness on benchmark datasets such as CIFAR10, Imagenet and so forth. However, this approach involves the cost of training or fine-tuning a large number of Subnets drawn from the parent Supernet. The NAS methodology in \cite{guo2020meets} is completely hardware-agnostic and hence, such models on crossbars with non-idealities will suffer huge degradation in their adversarial performance. 

Concerning NAS for analog CIM platforms, a recent work called NAX \cite{negi2021nax} explores the design space to determine appropriate kernel and corresponding crossbar sizes for each DNN layer to achieve optimal trade-offs between hardware-efficiency and application accuracy in presence of non-idealities. The DNN architectures derived using NAX have \textit{heterogeneous} crossbar sizes across different convolutional layers to achieve optimal trade-off between EDAP and application accuracy in presence of non-idealities. However, such heterogeneous architectures are not practical and are challenging to manufacture, since different-sized crossbars will require peripheral circuits to be custom designed \cite{zhu2018mixed, krishnan2020interconnect}. Moreover, no prior work in connection with NAS for analog crossbars has considered adversarial robustness as a key goal for optimization. 

There have been several previous works such as \cite{bhattacharjee2021neat, bhattacharjee2020switchx, bhattacharjee2021efficiency, roy2020robustness} that have used non-ideality driven techniques to improve the adversarial robustness of pretrained DNN models on memristive crossbar arrays. Recent works such as \textit{SwitchX} \cite{bhattacharjee2020switchx} and NEAT \cite{bhattacharjee2021neat} propose mapping DNNs onto crossbars in a manner that increases the proportion of low conductance synapses in each crossbar array. This helps in non-ideality mitigation, whereby the adversarial robustness of the DNN models is significantly improved on hardware. But, all these approaches have considered a fixed DNN model and do not propose architectural modifications to mitigate the effects of crossbar noise or adversarial noise. In contrast, our \textit{XploreNAS} approach is driven towards architecture search, to obtain an optimal trade-off between crossbar-based adversarial accuracy and EDAP.

\section{Background}
\label{sec:background}

\subsection{Adversarial Attacks}
\label{sec:adv}

Adversarial attacks have been shown to degrade a DNN's performance by introducing structured but small perturbations to the clean inputs, leading to high confidence misclassification. In this work, we use one of the strongest known gradient-based adversarial attacks in literature, \textbf{Projected Gradient Descent (PGD)}. The PGD attack, shown in eq. (\ref{pgd}), is an iterative attack over $n$ steps. In each step $i$, perturbations of strength $\alpha$ are added to $x_{adv}^{i-1}$. Note that $x_{adv}^{0}$ is created by adding random noise to the clean inputs $x$, $\theta$ denotes the DNN model's weight parameters, $y_{true}$ denotes the correct prediction labels for the inputs $x$ and $\mathcal{L}$ the cross-entropy loss function. Additionally, for each step, $x_{adv}^{i}$ is projected on a \textit{Norm ball} \cite{madry2017towards}, of radius $\epsilon$. In other words, we ensure that the maximum pixel difference between the clean and adversarial inputs is $\epsilon$. 
\begin{equation}
    x_{adv} = \sum_{i=1}^{n} x_{adv}^{i-1}+\alpha* ~sign(\nabla_x\mathcal{L}(\theta, x, y_{true})).
    \label{pgd}
\end{equation}
In this work, we will use the notation PGD-$n$ to denote a PGD attack iterated over $n$ steps. Further, we consider white-box attacks in our experiments where, the attacker is assumed to have full knowledge of the target model and dataset.

\subsection{Memristive Crossbars and Non-Idealities}
\label{sec:xbar_ni}

\begin{wrapfigure}{l}{0.25\textwidth}
\includegraphics[width=0.25\textwidth]{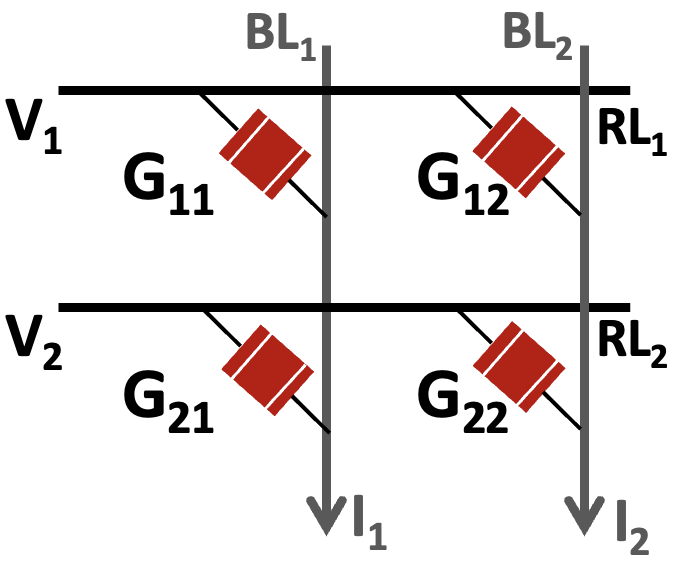}
  
\caption{A 2$\times$2 crossbar array with input voltages $V_{i}$, NVM devices bearing synaptic conductances $G_{ij}$ and output currents $I_{j}$.}
\label{xbar_img}
\vspace{-4mm}
\end{wrapfigure}

Memristive crossbars consist of 2D arrays of Non-Volatile-Memory (NVM) devices, Digital-to-Analog Converters (DACs), Analog-to-Digital Converters (ADCs) and a write circuit. The synaptic devices at the cross-points are programmed to a particular value of conductance (between $G_{MIN}$ and $G_{MAX}$) during inference. The MAC operations are performed by converting the digital inputs to the DNN into analog voltages on the Read lines (RLs) using DACs, and sensing the output current flowing through the Bitlines (BLs) using the ADCs \cite{jain2020rxnn, hu2016dot, bhattacharjee2021neat}. In other words, the activations of the DNNs are fed in as analog voltages $V_i$ to each row and weights are programmed as synaptic device conductances ($G_{ij}$) at the cross-points as shown in Fig. \ref{xbar_img}. For an ideal crossbar array, during inference, the voltages interact with the device conductances and produce a current (governed by Ohm's Law). Consequently, by Kirchoff's current law, the net output current sensed at each column $j$ is the sum of currents through each device, \textit{i.e.} $I_{j(ideal)} = \Sigma_{i}^{}{G_{ij} * V_i}$. We term the matrix $G_{ideal}$ as the collection of all $G_{ij}$'s for a crossbar. 

In reality, the analog nature of the computation leads to various hardware noise or non-idealities, such as, interconnect parasitic resistances and NVM device-level variations \cite{jain2020rxnn, chakraborty2020geniex, moitra2022spikesim}. This results in a $G_{non-ideal}$ matrix, with each element $G_{ij}'$ incorporating the impact of the non-idealities. Consequently, the net output current sensed at each column $j$ in a non-ideal scenario becomes $I_{j(non-ideal)} = \Sigma_{i}^{}{G_{ij}' * V_i}$, which deviates from its ideal value. This manifests as accuracy degradation for DNNs mapped onto crossbars. Larger crossbars entail greater non-idealities, resulting in higher accuracy losses \cite{jain2020rxnn, chakraborty2020geniex, bhattacharjee2021efficiency}.

\section{Methodology of $XploreNAS$}
\label{sec:method}

Our NAS methodology is based on the conventional one-shot learning using DARTS \cite{bender2018understanding, liu2017learning}, that is adapted to include the impact of crossbar noise along with adversarial training \cite{madry2017towards}. The entire process of training an over-parameterized network (\textit{Supernet}) that includes a number of operations, and ultimately obtaining an optimal neural network configuration (\textit{Subnet}) is described as follows:

\subsection{Supernet Architecture}
\label{sec:supernet}

\begin{figure}[t]
    \centering
    \includegraphics[width=.75\linewidth]{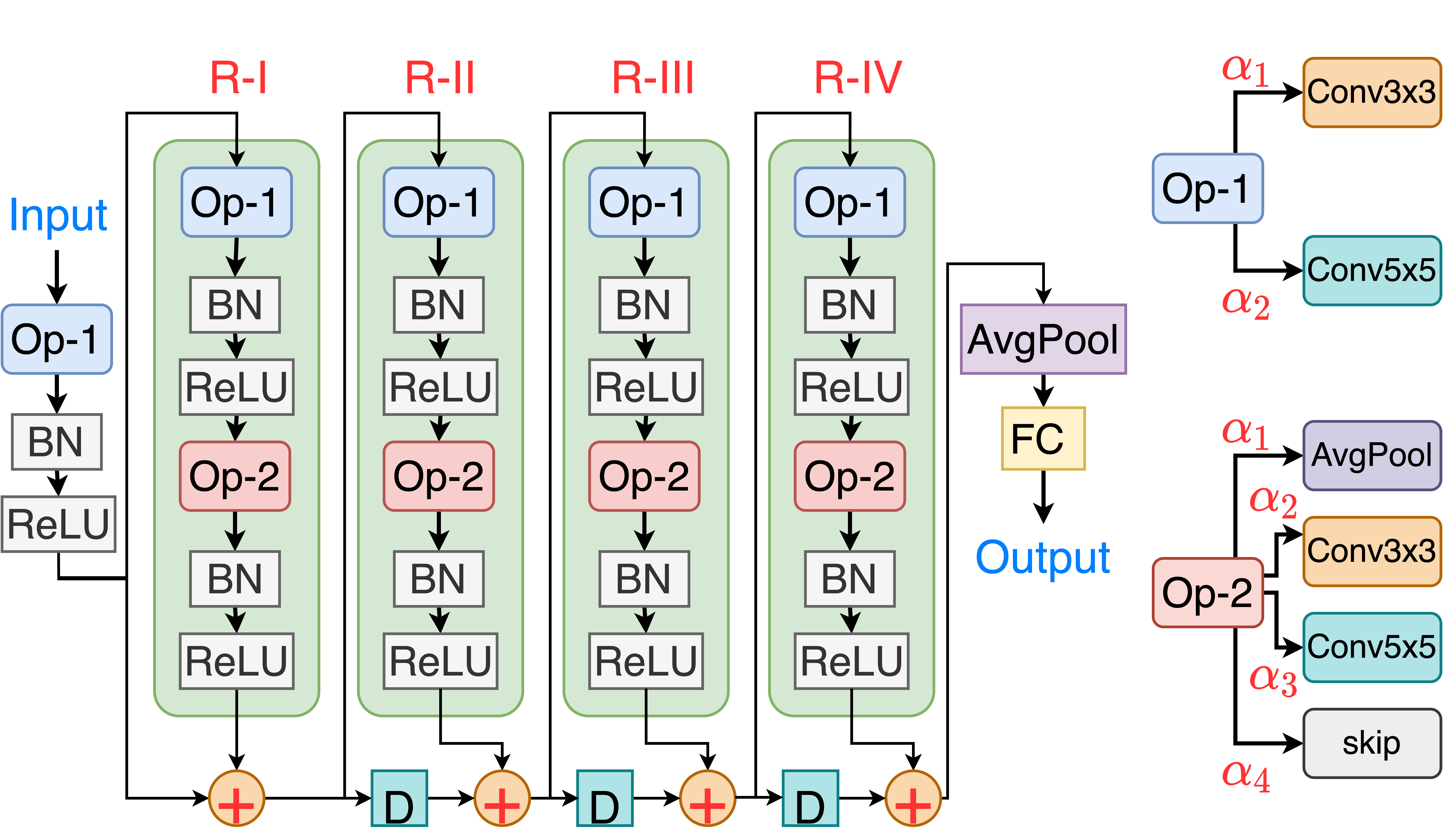}
   
    \caption{The Supernet architecture for \textit{XploreNAS} with the constituents of Op-1 and Op-2 operations indicated.}
    \label{supernet}
     \vspace{-4mm}
\end{figure}

Fig. \ref{supernet} presents our Supernet architecture. 
There are four residual blocks highlighted in green, from R-I to R-IV. At end of network after R-IV, there is an average pooling layer with a stride of 2 and kernel size of 3x3 followed by a full-connected classifier layer. In each residual block, we have an Op-1, Op-2 operation choice, each followed by a batchnorm layer and a ReLU function. Op-1 operation choice constitutes convolution operations with 3x3 and 5x5 kernel sizes (\textit{i.e.}, Conv3x3 and Conv5x5). The Op-2 operation choice constitutes of average pooling with a stride of 1 and kernel size of 3x3 (AvgPool), Conv3x3, Conv5x5 and a skip-connection (implying no operation). Note, a downsampling block (marked as $D$) is used at the end of residual blocks R-I, R-II and R-III since the feature sizes of the corresponding inputs and outputs of these residual blocks are unequal. The downsampling block includes a Conv3x3 operation (with a stride of 2) followed by a batchnorm layer. 

Each constituent operation choice in Op-1 and Op-2 is associated with a parameter $\alpha$ called the architecture parameter in the supernet. Let us suppose that a given Op-1 operation in the Supernet is associated with the architecture parameters $\alpha_{1}$ and $\alpha_{2}$ for the constituent Conv3x3 ($o_1$) and Conv5x5 ($o_2$) operations, respectively. If $p_{1}$ and $p_{2}$ are, respectively, the \textit{softmax} of $\alpha_{1}$ and $\alpha_{2}$, then the output at the end of the Op-1 operation ($m_{op-1}$) is computed using DARTS as follows:
\begin{equation} m_{op-1} = \sum_{j=1}^{2}p_j * o_j.
\label{eq:out-op1}
\end{equation}
Here, $p_j$ $\epsilon$ [0,1] pertaining to each constituent operation is referred to as a probability coefficient. Similarly, for a given Op-2 operation which has four constituent operations (AvgPool ($o_1$), Conv3x3 ($o_2$), Conv5x5 ($o_3$) and skip-connection ($o_4$)), the output ($m_{op-2}$) is computed as follows:

\begin{equation} m_{op-2} = \sum_{j=1}^{4}p_j * o_j.
\label{eq:out-op2}
\end{equation}


\subsection{Mapping a DNN onto Non-ideal Crossbars}

\label{sec:xbar_map}

\begin{figure}[t]
    \centering
    
    \includegraphics[width=.75\linewidth]{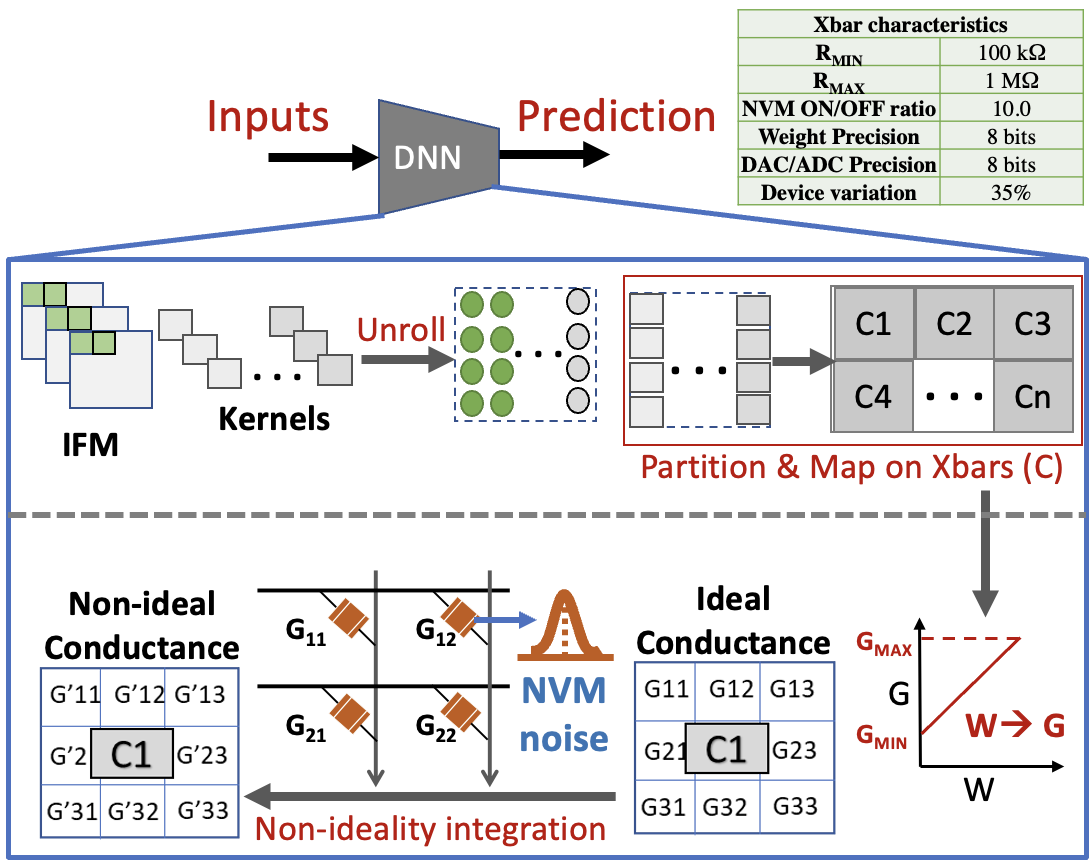}%
    \caption{Flow-diagram illustrating the mapping of DNN weight matrices on non-ideal memristive crossbars. The crossbar-related parameters are specified in the table enclosed.}
    \label{xbar_map}

\end{figure}

Here, we describe the manner in which weight matrices of different DNN layers are mapped onto crossbar arrays. The entire procedure is carried out in Python for better integration between the software model and the hardware mapping framework. In Fig. \ref{xbar_map}, we have a Python wrapper built to unroll each and every convolution operation in the software DNN model into MAC operations between input activation matrices and their corresponding weight matrices. For a given convolutional layer, its 4D weight matrix is reshaped to a 2D matrix $W$. 
The matrices obtained are then zero-padded (in case the dimensions of the 2D weight matrix $W$ are not exact multiples of the crossbar size $n \times n$) and then partitioned into multiple $n \times n$ crossbar arrays consisting of DNN weights at the synapses (modelled as conductances). Here, we assume that the NVM devices at the synapses in the crossbars can be programmed to a conductance level between $R_{MIN}=100k\Omega$ and $R_{MAX}=1M\Omega$, typical for ReRAM devices. For this range of conductance, the impact of crossbar non-idealities due to parasitic interconnect resistances is minimized \cite{roy2021txsim}. Thus, the NVM device-level non-linearities or stochasticity are the key players that impact the accuracy of DNNs deployed on such crossbars. Next, we add synaptic device variations to the weights mapped in the crossbar arrays. The device variations for each $n \times n$ crossbar are modelled using a Gaussian distribution with $\sigma/\mu$ of 35\% (assuming 8-bit precision of DNN weights when mapped to the NVM synapses in the crossbars) \cite{charan2020accurate, sun2019impact}. Note, this noise-profile is specific to the crossbar size under consideration. Finally, these noisy weights ($W_{noisy}$) are then integrated into the original Pytorch-based DNN model to facilitate crossbar-aware evaluation.

\subsection{Crossbar-aware Training of the Supernet}
\label{sec:xnas}

\begin{algorithm}
    \caption{Crossbar-aware Supernet Training}

\RaggedRight \textbf{Input}: Supernet $S$ with weights $W$ and architecture parameters $\alpha$, total number of epochs $I$, batch-size $B$, Training and Validation$^*$ sets with inputs and targets $\{x_i,y_i\}$\\
  
\textbf{Output}: Optimal Subnet architecture $\Phi$
  
  \begin{algorithmic}[1]
    \For{$u = 1$ to $I$}
        \State{// \textbf{Phase-1 Training}}
        
        \State{Randomly sample a batch $B$ of clean inputs from the Training set $\{x_i,y_i\}_{i=1}^B$}

        \State{Use $\{x_i,y_i\}_{i=1}^B$ to do one-step training of $W$ using Adam Optimizer}
        
        \State{// \textbf{Phase-2 Training}}
        
        \State{Partition $W$ into multiple crossbar-arrays of given size $n\times n$}
        
        \State{Add crossbar-specific noise ($\tilde{n}$) to the weights in the arrays, \textit{i.e.} $W \leftarrow W+\tilde{n}$}
        
        \For{$v = 1$ to $N$} // Assuming there are $N$ batches of $\{x_i,y_i\}_{i=1}^B$ in the Validation set
            
            \State{Generate adversarial inputs $\{x_i^{adv},y_i\}_{i=1}^B$ from the clean inputs $\{x_i,y_i\}_{i=1}^B$ of the Validation set by randomly creating white-box PGD-7 or PGD-20 attack}
            
            \State{Use $\{x_i^{adv},y_i\}_{i=1}^B$ to train the architecture parameters $\alpha$ using Adam Optimizer}
            
        \EndFor
    \EndFor\\
    The Subnet $\Phi$ is derived from the Supernet $S$ using the trained architecture parameters $\alpha$ (see Section \ref{sec:subnet_derive}).
     
  \end{algorithmic}
       \label{algorithm:supernet-train}

   \scriptsize$^*$Note, the Validation dataset is a subset of the original Training dataset and is created by randomly sampling 5000 images from the Training set.\normalsize
\end{algorithm}

Having proposed the Supernet architecture, we now describe the methodology adopted for carrying out crossbar-aware NAS (\textit{XploreNAS}). One epoch of training the Supernet architecture occurs in two phases (see Algorithm \ref{algorithm:supernet-train}) as follows:

\subsubsection{Phase-1: Training the weight parameters}

A batch of clean inputs with a large batch-size (1000 in this work) is randomly sampled from the training dataset and forwarded through the Supernet once. Thereafter, the weight parameters for all the layers in the Supernet are updated using backpropagation. Note, during Phase-1, we do not include any hardware-related parameter during the weight update. Also, the architectural parameters ($\alpha$) are unaffected in Phase-1.

\subsubsection{Phase-2: Training the architecture parameters with hardware-awareness}

After Phase-1, the weight matrices for the different convolutional layers in the Supernet are partitioned into numerous $n \times n$ non-ideal crossbar arrays and integrated with crossbar-level noise using the methodology described in Section \ref{sec:xbar_map} as shown in Fig. \ref{xbar_map}. 
Note that in Phase-2, the noisy weights are frozen and only the architecture parameters ($\alpha$) are trained using backpropagation. 
The Supernet integrated with crossbar-level noise (for a given crossbar size) is subjected to ensemble of adversarial images from the validation dataset (a subset of the original training dataset) sent in batches. For a given batch, adversarial images are generated either using PGD-7 or PGD-20 attack randomly. The architecture parameters are updated to assign higher probability coefficients to a series of operations or paths through the Supernet that are more resilient to the impact of crossbar noise as well as adversarial noise. We repeat the above steps and train the Supernet for a few epochs, each epoch consisting of Phase-1 followed by Phase-2 training. 

\subsection{Deriving the optimal Subnet from the trained Supernet}

\label{sec:subnet_derive}
Finally, based on the trained architecture parameters, we sample the optimal Subnet from the over-parameterized Supernet by modifying eq. (\ref{eq:out-op1}) \& (\ref{eq:out-op2}) based on the following criteria: 

\begin{equation} m_{op-1~or~op-2}  = \sum_{\forall j}^{}g_j * o_j,
\label{eq:out-final}
\end{equation}

where, 

\begin{equation}
    g_j =
\begin{cases}
 1 ,   & p_j >  th \\
    0 , & p_j \leq th.
\end{cases}
\label{eq:gates}
\end{equation}

In other words, if probability coefficient for a constituent operation of Op-1 or Op-2 is greater than a threshold $th$, we retain the operation in the computation of the output at the end of Op-1 or Op-2. Otherwise, the operation is skipped. In this way, we derive our single Subnet with optimal number of operations from the trained Supernet architecture. In this work, the value of $th$ is heuristically chosen to be 0.2, for the Subnet to achieve substantially higher robustness against the baselines. 

\subsection{Fine-tuning the Subnet}
\label{sec:fine-tune}

Having obtained the Subnet architecture, we train the weights of the model using backpropagation by feeding an ensemble of clean and adversarial images (from the training dataset). 
Note, similar to Phase-2, we include crossbar-specific noise in the weights of the Subnet during the fine-tuning phase. Further, the adversarial images used in this stage can be generated either using white-box PGD-7 or PGD-20 attack. 

\subsection{Deriving three types of Subnet models}

In this work, we derive three types of Subnet models, namely:

\textbf{Model Xbar}: This is the standard scenario wherein, the noise-profile specific to a given crossbar size is integrated with the model weights during Phase-2 of Supernet training and Subnet fine-tuning. 

\textbf{Model Xbar\_Ar}: Here, in addition to adding crossbar-specific noise-profile to weights in the Supernet, we also regularize the loss function during Phase-2 so as to search for a Subnet architecture that is optimized to have minimal resource-utilization on memristive crossbars. 
The simplest way to accomplish this is to optimize the architectures with respect to crossbar-area consumption. We follow the differentiable approach proposed in \cite{cai2018proxylessnas} to optimize crossbar-specific area utilization during Phase-2. Let $\phi$ denote the hardware parameter (area) that needs to be optimized. Thus, for Op-1 or Op-2 constituting multiple paths or operations each with a given probability coefficient, we have the following expression for the expected value of $\phi$, \textit{i.e.} $E[\phi]$:
\begin{equation} E[\phi]_{op-1~or~op-2}  = \sum_{\forall j}^{}p_j * \phi_j,
\label{eq:exp-el}
\end{equation}

where, $\phi_j$ denotes the area estimate for the $j^{th}$ constituent operation in Op-1 or Op-2, and all the possible values for $\phi_j$ for a given crossbar size can be found in a lookup table
containing the area estimates for all the operations in the
search space. Now, by summing $E[\phi]_{op-1~or~op-2}$ over all the Op-1 and Op-2 operations in the Supernet model, we get the cumulative expected value of $\phi$ for the Supernet (denoted as $E[\phi]_{total}$), that needs to be minimized. Hence, the loss function in Phase-2 ($\Lagr_{Phase-2}$) is written as follows:

\begin{equation} \Lagr_{Phase-2} = \Lagr_{CE} + \lambda*E[\phi]_{total}. 
\label{eq:loss_el}
\end{equation}

Here, $\Lagr_{CE}$ denotes the cross-entropy loss and $\lambda$ is a hyperparameter that control the relative importance given to $E[\phi]_{total}$ with respect to $\Lagr_{CE}$.

\textbf{Model MultiXbar}: Here, our objective is to search for a Subnet architecture that is not crossbar-specific, rather robust across multiple crossbar sizes. Let us suppose a scenario wherein we want to obtain a Subnet that is robust against hardware-noise pertaining to 32$\times$32, 64$\times$64 and 128$\times$128 crossbars. Thus, during Phase-2, for a batch of adversarial inputs fed into the Supernet, we first add the noise-profile for 32$\times$32 crossbars to the weights, freeze them, update the architecture parameters and restore the noise-free weights. Again, for the same batch of inputs, we add the noise-profile for 64$\times$64 crossbars to the weights, freeze them, update the architecture parameters and restore the noise-free weights. Finally, the same process is repeated for 128$\times$128 crossbars. In this manner, we ultimately arrive at a Subnet architecture that is resilient to the noise-profiles of multiple crossbar sizes. Thereafter, at the end of Phase-2 training, we derive the optimal Subnet (eq. \ref{eq:out-final} \& \ref{eq:gates}) and then, fine-tune (see Section \ref{sec:fine-tune}) our Subnet using an ensemble of noise-profiles pertaining to different crossbar sizes.

\section{Metrics for Assessment}

\label{sec:metrics}

In this section, we define the metrics that are used to evaluate the Subnets as well as our baseline models on memristive crossbar arrays. These are as follows:

\textbf{Adversarial accuracy on crossbars:} It denotes the PGD-$n$ accuracy of the trained Subnets and baselines on non-ideal crossbar arrays during inference. Higher the value of adversarial accuracy, better the robustness of the DNN deployed on crossbars.

\textbf{Hardware-efficiency using EDAP:} We compute the overall EDAP for a DNN during inference on a memristive crossbar-based hardware platform using the Neurosim tool. Neurosim \cite{chen2018neurosim} is a Python-based hardware-evaluation platform that performs a holistic energy-latency-area evaluation of analog crossbar-based DNN accelerators. The EDAP evaluations using Neurosim include contributions of the dot-product processing engines (\textit{i.e.,} the crossbars) as well as the peripheral circuits (DACs, ADCs, buffers and so forth) and on-chip interconnects. Note in this work, all EDAPs are shown in the units of $mJ.ms.mm^2$. Higher the EDAP, better the hardware-efficiency of the DNN model. We calibrate the Neurosim environment for ReRAM crossbar arrays with specifications listed in the table inscribed within Fig. \ref{xbar_map}. 

\textbf{Average crossbar-underutilization:} As discussed in Section \ref{sec:xbar_map}, if the dimensions of the 2D weight matrix of a DNN layer are not exact multiples of the crossbar size, then the weight matrix is zero-padded and partitioned into multiple crossbar arrays. This zero-padding effectively results in the underutilization of certain crossbar arrays, \textit{i.e.}, additional hardware area, energy, latency and leakage power are expended in the computation and processing of certain non-useful dot-products. Let us assume that the dimensions of a 2D weight matrix ($in\_ch*k^2$, $out\_ch$) are not multiples of crossbar size $n \times n$. Then, the amount of zero-padding along the rows is given by $r_{pad} = (in\_ch*k^2)+(in\_ch*k^2)\%n$, where the operator \% denotes the remainder operation. Likewise, the amount of zero-padding along the columns is given by $c_{pad} = out\_ch+out\_ch\%n$. Thus, we define the value of crossbar-underutilization for the given 2D weight matrix as follows:

\begin{equation} 
\begin{aligned}
    crossbar-underutilization = \frac{in\_ch*k^2*c_{pad}+out\_ch*r_{pad}+r_{pad}*c_{pad}}{(in\_ch*k^2+r_{pad})*(out\_ch+c_{pad})}*100.
\end{aligned}
\label{eq:xbar_under}
\end{equation}

If there are $l$ convolutional layers in a DNN whose weight matrices are to be zero-padded, then average crossbar-underutilization for the DNN is defined as the mean of the values of crossbar-underutilization for the $l$ layers.

\section{Experiments and Results}
\label{sec:expt}

\begin{figure}[htbp]
    \centering
    \subfloat[]{
    \includegraphics[width=0.4\linewidth]{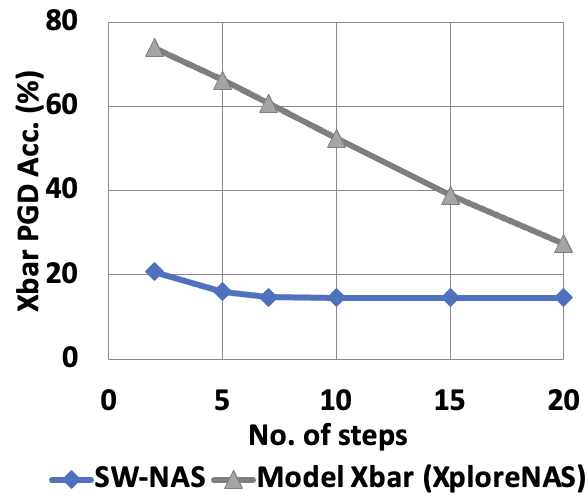}%
    } 
    \subfloat[]{
    \includegraphics[width=0.4\linewidth]{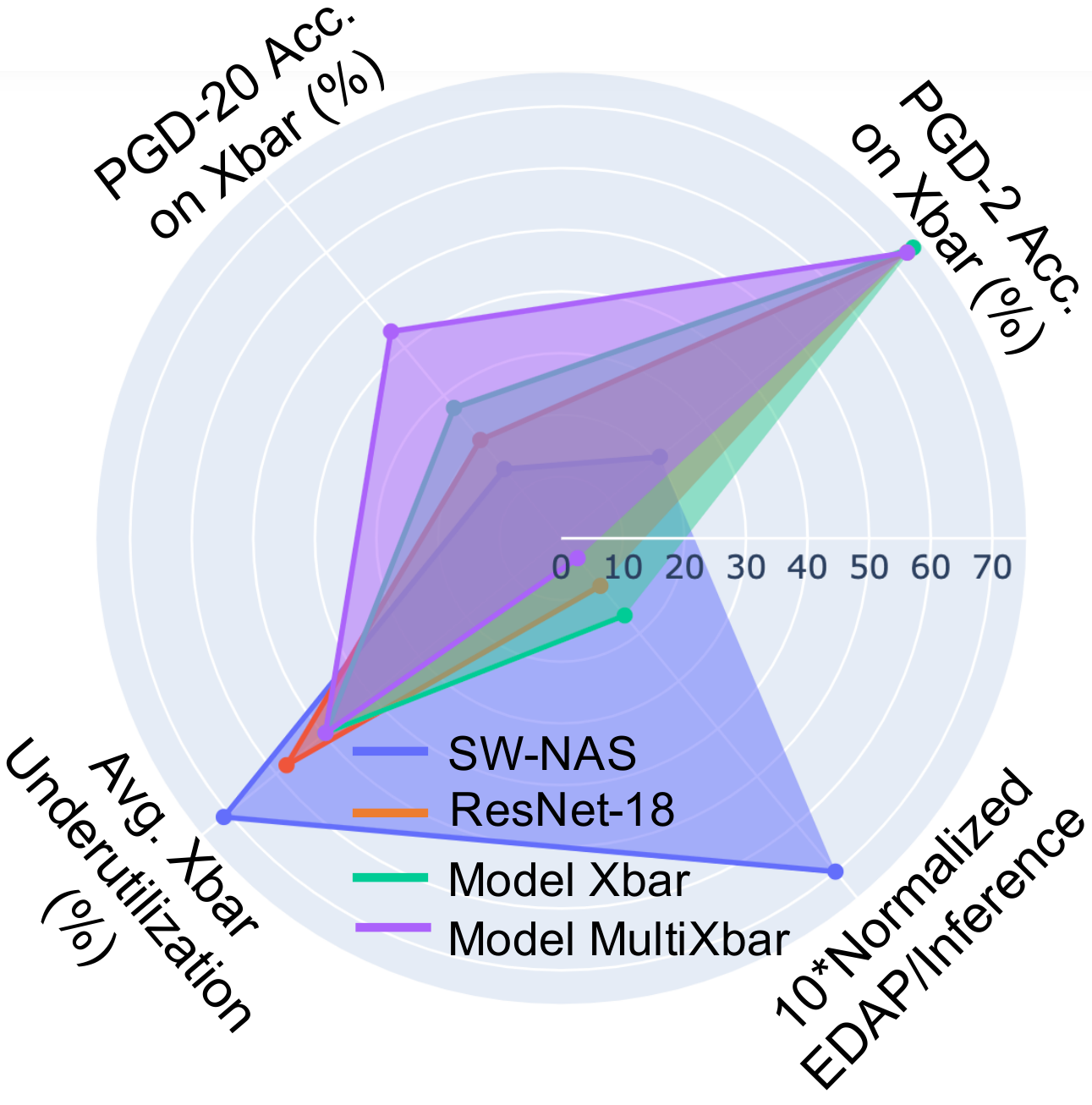}
    }
    \caption{(a) Plot of adversarial CIFAR10 accuracies of Model Xbar and SW-NAS model \cite{guo2020meets} on 64$\times$64 crossbars with respect to the number of steps in PGD-$n$ attack. (b) A radar-chart comparing \textit{XploreNAS} generated Models Xbar \& MultiXbar, baseline ResNet-18 and SW-NAS models on 64$\times$64 crossbars for adversarial robustness on crossbars and hardware-efficiency using CIFAR10 dataset. }
    \label{robnet_comps}
    \vspace{-4mm}
\end{figure}

We conduct our experiments with benchmark datasets- SVHN \cite{netzer2011reading}, CIFAR10 \& CIFAR100 \cite{cifar}, using Pytorch. We construct the validation set by selecting 5,000 images randomly from the overall training dataset. The remainder of images constitute our training dataset. Note, the test dataset used to evaluate the inference accuracy of the DNN models on hardware is completely different from the training and validation sets and is never used during the Supernet training or Subnet fine-tuning stages. We perform the two-phase crossbar-aware training of the Supernet for 60 epochs using  Algorithm \ref{algorithm:supernet-train}. The crossbar sizes considered in this work include 32$\times$32, 64$\times$64 and 128$\times$128. We generate Model XBar, Model Xbar\_Ar and Model MultiXbar using \textit{XploreNAS}. These Subnets are fine-tuned using Adam Optimizer for 40 epochs (see Section \ref{sec:subnet_derive}) and used for inference on crossbars. Unless otherwise stated, the crossbar-related parameters used for training and inference are specified in the table inscribed within Fig. \ref{supernet}(b). Note, all PGD-$n$ adversarial attacks are white-box in nature with $\alpha=2/255$ and $\epsilon=8/255$. Our baseline is a ResNet-18 architecture adversarially trained for 40 epochs using Adam Optimizer with an ensemble of PGD-7 and PGD-20 inputs in presence of crossbar-specific noise for a given crossbar size. The crossbar-aware clean accuracies (\textit{i.e.}, inference accuracy in absence of adversarial attack) of the \textit{XploreNAS}-derived Subnets and ResNet-18 baselines on 64$\times$64 crossbars have been listed in Table \ref{tab:my-table}. It should be noted that all of these models have been adversarially trained in presence crossbar-specific noise profiles. The rigorous adversarial training in presence of an ensemble of noise-profiles pertaining to different crossbar sizes for Model MultiXbar limits its clean accuracy compared to the other \textit{XploreNAS}-derived Subnets, although we will see in the following sections that Model MultiXbar achieves the best hardware-aware adversarial robustness across multiple crossbar sizes with competitive EDAP benefits.

\begin{wraptable}{l}{4.5cm}
\caption{Table showing the hardware-aware clean accuracies of the \textit{XploreNAS}-derived Subnets and ResNet-18 baseline on 64$\times$64 crossbars for CIFAR10 and CIFAR100 datasets.}
\label{tab:my-table}
\resizebox{0.3\textwidth}{!}{%
\begin{tabular}{|c|c|c|}
\hline
\textbf{Dataset} & \textbf{Model} & \textbf{\begin{tabular}[c]{@{}c@{}}Xbar Clean \\ Accuracy (\%)\end{tabular}} \\ \hline
\multirow{4}{*}{CIFAR10}  & \begin{tabular}[c]{@{}c@{}}ResNet-18\\ (Baseline)\end{tabular} & 80.06 \\ \cline{2-3} 
                          & Model Xbar                                                     & 84.32 \\ \cline{2-3} 
                          & Model Xbar\_Ar                                                 & 83.87 \\ \cline{2-3} 
                          & Model MultiXbar                                                & 81.03 \\ \hline
\multirow{4}{*}{CIFAR100} & \begin{tabular}[c]{@{}c@{}}ResNet-18\\ (Baseline)\end{tabular} & 51.36 \\ \cline{2-3} 
                          & Model Xbar                                                     & 55.82 \\ \cline{2-3} 
                          & Model Xbar\_Ar                                                 & 54.62 \\ \cline{2-3} 
                          & Model MultiXbar                                                & 51.89 \\ \hline
\end{tabular}%
}
\end{wraptable}

Now, we present Fig. \ref{robnet_comps} that helps us understand the usefulness of our proposed hardware-aware \textit{XploreNAS} approach. The results are for 64$\times$64 crossbars on CIFAR10 dataset. In Fig. \ref{robnet_comps}(a), we find that Model Xbar achieves a high degree of adversarial robustness on hardware compared to a purely software-based adversarially trained DNN model derived via NAS as per \cite{guo2020meets}, when inferred on crossbars with non-idealities. Note, the purely software-based model (referred to as SW-NAS) achieves state-of-the-art adversarial accuracy ($\sim52-56\%$) on software using CIFAR10 dataset against white-box PGD attacks. However on crossbars for weaker attack (PGD-2), the adversarial accuracy of SW-NAS witnesses a huge drop to $\sim20\%$, while our Model Xbar achieves $\sim74\%$ adversarial accuracy. For stronger attack (PGD-20), the accuracy of SW-NAS is at $\sim14\%$, while Model Xbar still achieves $\sim13\%$ higher accuracy than SW-NAS. 

Fig. \ref{robnet_comps}(b) presents a radar chart that helps us understand the usefulness of our proposed NAS approach across multiple dimensions. Here, we look at the EDAP per inference estimated using the Neurosim tool, normalized with respect to the baseline ResNet-18 model. We find that the SW-NAS model has $\sim4.5\times$ higher EDAP than Model Xbar, while the ResNet-18 baseline is slightly more hardware-efficient than Model Xbar (which is not as optimized as its Model Xbar\_Ar counterpart). Although Model MultiXbar achieves marginally higher clean accuracy with regard to the ResNet-18 baseline as shown in Table \ref{tab:my-table}, it provides the best trade-off between adversarial accuracy and EDAP on hardware, achieving $\sim30\%$ higher accuracy than SW-NAS for stronger PGD attack (PGD-20) and has $\sim2\times$ lower EDAP than the ResNet-18 baseline. Another dimension to look at in Fig. \ref{robnet_comps}(b) is the average crossbar-underutilization. As discussed in Section \ref{sec:metrics}, higher the crossbar-underutilization, greater is the hardware energy and area expended on the processing of non-useful dot-product computations (or MAC operations). SW-NAS, being optimized in a hardware-agnostic manner, has a layerwise architecture that is not crossbar-friendly and leads to a huge average crossbar-underutilization of $\sim71\%$. For our Model Xbar, the crossbar-underutilization gets reduced to $\sim49\%$ and is restricted only to the Op-1 operation prior to the R-I residual-block of our Supernet. This is attributed to our crossbar-centric choice of kernel sizes and input/output channels in the convolutional layers of the Supernet model.

\label{sec:acc_result}

\begin{figure*}[t]
    \centering
    
    \includegraphics[width=\linewidth]{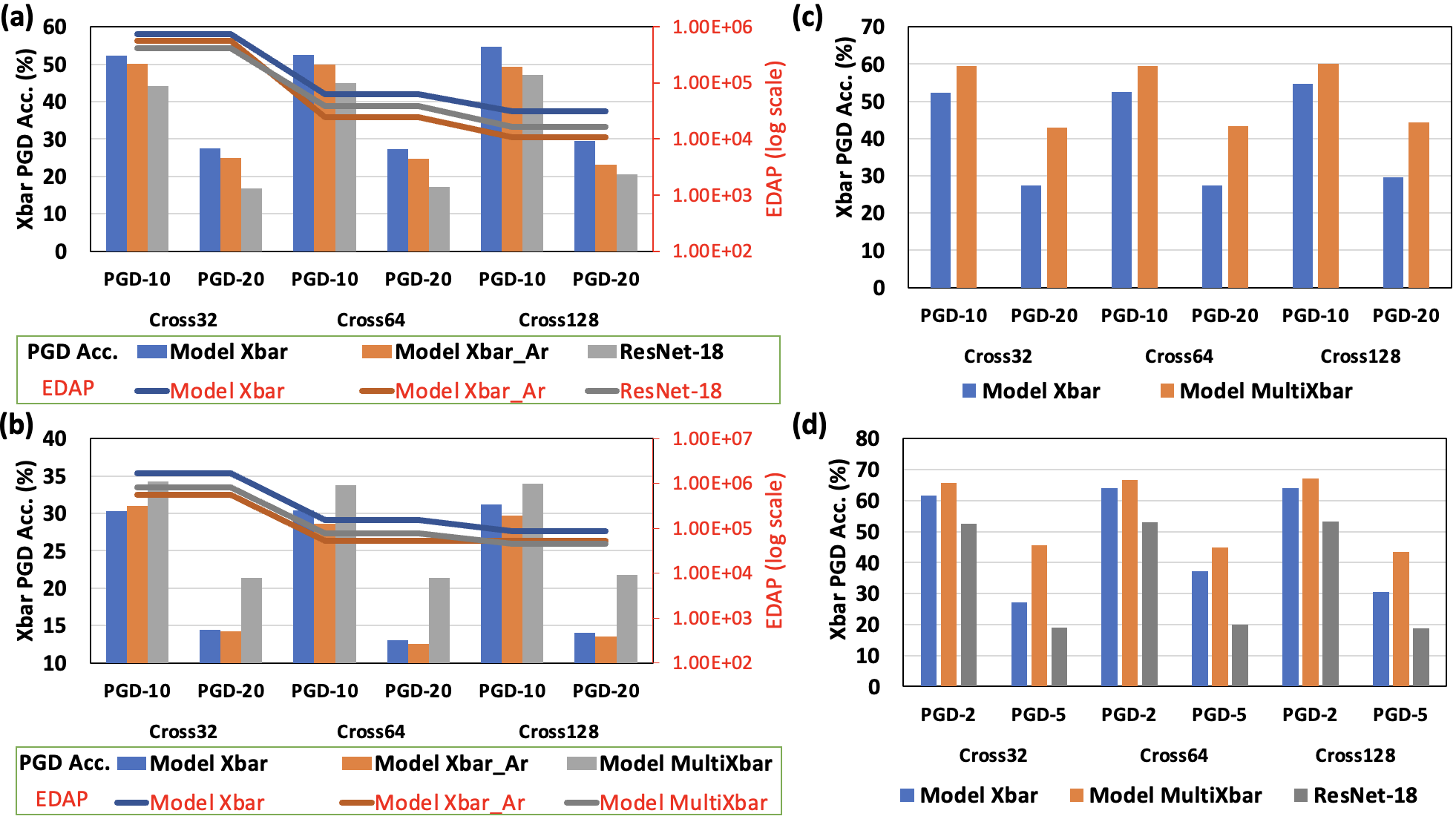}%
    
    \caption{Plot of adversarial accuracies (primary y-axis) and EDAPs (secondary y-axis) on crossbars for: (a) Model Xbar, Model Xbar\_Ar and ResNet-18 baseline using CIFAR10 dataset, (b) Model Xbar, Model Xbar\_Ar and ResNet-18 baseline using CIFAR100 dataset. Plot of adversarial accuracies on crossbars for: (c) Model Xbar and Model MultiXbar using CIFAR10 dataset, (d) Model Xbar, Model MultiXbar and ResNet-18 baseline using SVHN dataset.}
    \label{cifar_acc_edap}

\end{figure*}

\subsection{Analysis of adversarial accuracy of Subnets}

The results in Fig.\ref{cifar_acc_edap}(a) \& (c) are for CIFAR10 dataset, Fig.\ref{cifar_acc_edap}(b) is for CIFAR100 dataset and Fig.\ref{cifar_acc_edap}(d) for SVHN dataset. In Fig.\ref{cifar_acc_edap}(a), we find that Model Xbar corresponding to crossbar sizes ranging from 32$\times$32 to 128$\times$128 are more adversarially robust ($\sim2-4\%$) than their Model Xbar\_Ar counterparts. This is because Model Xbar is purely optimized to maximise adversarial accuracy on noisy crossbars, while Model Xbar\_Ar is optimized to yield crossbar-efficient architectures for hardware-constrained scenarios in addition to maximising adversarial robustness. However, both Model Xbar and Model Xbar\_Ar outperform their corresponding ResNet-18 baselines in terms of adversarial robustness across all the crossbar sizes considered. 

Furthermore, in Fig.\ref{cifar_acc_edap}(c) we find that Model MultiXbar optimized for crossbar noise pertaining to sizes ranging from 32$\times$32 to 128$\times$128 attains significantly higher robustness with respect to the Model Xbar counterparts, specifically for stronger PGD-20 attacks ($\sim15-16\%$ better robustness). This is because for each training epoch during Subnet fine-tuning in Model MultiXbar, we update weights thrice (corresponding to three different crossbar noise-profiles) for a given batch of inputs. Thus, the Subnet attains inherently higher hardware noise-aware adversarial robustness than the three Model Xbar architectures each pertaining to a specific crossbar size. This analysis also shows that \textit{XploreNAS} can derive a Subnet architecture that can be inferred on crossbar arrays of varying sizes with high degree of adversarial resilience across weak and strong PGD attacks. Similar results can be seen for the Subnets trained with CIFAR100 dataset as depicted in Fig.\ref{cifar_acc_edap}(b) with Model MultiXbar outperforming the corresponding Model Xbar by a margin of $\sim4-8\%$ across the PGD-10 and PGD-20 adversarial attacks. For the plot in Fig. \ref{cifar_acc_edap}(d) for SVHN dataset, we show results for weak attacks (PGD-2 and PGD-5) since stronger attacks using a simple dataset like SVHN leads to random test accuracy for both NAS-derived and ResNet-18 models. We see similar trends as that of CIFAR10 dataset. Note, here Model Xbar and Model Xbar\_Ar comprise of three corresponding Subnets/DNNs for three different crossbar sizes, while, Model MultiXbar is a single Subnet/DNN optimized for all three crossbar sizes.

\subsection{Analysis of EDAP results of Subnets}
\label{sec:el_result}

In this section, we benchmark our Subnet architectures derived using \textit{XploreNAS} for overall hardware-efficiency using the Neurosim tool. In Fig. \ref{cifar_acc_edap}(a), we plot the EDAP per inference for various DNNs on the logarithmic scale with respect to the crossbar size. The trends pertain to CIFAR10 dataset. We find that although Model Xbar outperforms the corresponding baseline ResNet-18 model in terms of adversarial robustness on crossbars, it expends a higher EDAP on hardware ($\sim1.62\times$ higher on 64$\times$64 crossbars). This scenario is altered on considering Model Xbar\_Ar, which is optimized to yield better hardware-efficiency in addition to adversarial robustness. On 64$\times$64 (128$\times$128) crossbars, Model Xbar\_Ar outperforms the corresponding ResNet-18 baseline by achieving $\sim1.60\times$ ($\sim1.53\times$) lower EDAP on hardware. In fact, the EDAP of Model Xbar\_Ar on 64$\times$64 crossbars is lower than Model Xbar on 128$\times$128 crossbars.

The EDAP trends in Fig. \ref{cifar_acc_edap}(b) pertain to CIFAR100 dataset. The plot corroborates that Model MultiXbar, that is more robust than the corresponding Model Xbar, is also inherently more hardware-efficient without any additional optimizations during training. This result implies that \textit{XploreNAS} inherently performs a performance vs. EDAP trade-off when searching for a Subnet in Phase-2 training that is amenable for all three crossbar size-specific noise profiles (Model MultiXbar scenario). On 128$\times$128 crossbars, Model MultiXbar even achieves $\sim1.16\times$ lower EDAP than the corresponding Model Xbar\_Ar, which is specifically optimized to better hardware-efficiency.

\subsection{How do the searched architectures differ?}

\begin{table*}[htbp]
\caption{Table showing the architectures of the different \textit{XploreNAS}-derived Subnets}
\label{tab:architectures}
\resizebox{\columnwidth}{!}{%
\begin{tabular}{|c|c|c|c|}
\hline
\textbf{Dataset} &
  \textbf{Crossbar size} &
  \textbf{Subnet type} &
  \textbf{Architecture (\textit{Select}   operations arranged sequentially)} \\ \hline
\multirow{7}{*}{\textbf{CIFAR10}} &
  \textbf{32} &
  \multirow{3}{*}{\textbf{Model Xbar}} &
  Conv3x3, Conv5x5 $\rightarrow$ \textcolor{blue}{Conv3x3, Conv5x5} $\rightarrow$   AvgPool $\rightarrow$ Conv3x3, Conv5x5 $\rightarrow$ AvgPool $\rightarrow$ \textcolor{blue}{Conv5x5}   $\rightarrow$ AvgPool, skip $\rightarrow$ \textcolor{blue}{Conv5x5} $\rightarrow$ \textcolor{red}{Conv3x3} \\ \cline{2-2} \cline{4-4} 
 &
  \textbf{64} &
   &
  Conv3x3, Conv5x5 $\rightarrow$ \textcolor{blue}{Conv5x5} $\rightarrow$   AvgPool $\rightarrow$ Conv3x3, Conv5x5 $\rightarrow$ AvgPool $\rightarrow$ \textcolor{blue}{Conv3x3}   $\rightarrow$ AvgPool, skip $\rightarrow$ \textcolor{blue}{Conv3x3} $\rightarrow$ \textcolor{red}{Conv3x3} \\ \cline{2-2} \cline{4-4} 
 &
  \textbf{128} &
   &
  Conv5x5 $\rightarrow$ \textcolor{magenta}{\textbf{Conv3x3, Conv5x5}} $\rightarrow$   AvgPool $\rightarrow$ Conv3x3 $\rightarrow$ AvgPool $\rightarrow$ \textcolor{magenta}{\textbf{Conv5x5}}   $\rightarrow$ AvgPool, skip $\rightarrow$ \textcolor{magenta}{\textbf{Conv5x5}} $\rightarrow$ \textcolor{red}{Conv3x3} \\ \cline{2-4} 
 &
  \textbf{32} &
  \multirow{3}{*}{\textbf{Model Xbar\_Ar}} &
  Conv3x3, Conv5x5 $\rightarrow$ Conv3x3, Conv5x5 $\rightarrow$   AvgPool $\rightarrow$ Conv3x3, Conv5x5 $\rightarrow$ AvgPool $\rightarrow$ Conv3x3   $\rightarrow$ AvgPool, skip $\rightarrow$ Conv3x3, Conv5x5 $\rightarrow$ \textcolor{red}{skip} \\ \cline{2-2} \cline{4-4} 
 &
  \textbf{64} &
   &
  Conv3x3, Conv5x5 $\rightarrow$ Conv3x3, Conv5x5 $\rightarrow$   AvgPool $\rightarrow$ Conv3x3 $\rightarrow$ AvgPool $\rightarrow$ Conv3x3   $\rightarrow$ AvgPool, skip $\rightarrow$ Conv3x3 $\rightarrow$ \textcolor{red}{skip} \\ \cline{2-2} \cline{4-4} 
 &
  \textbf{128} &
   &
  Conv3x3, Conv5x5 $\rightarrow$ Conv3x3, Conv5x5 $\rightarrow$   AvgPool $\rightarrow$ Conv5x5 $\rightarrow$ AvgPool $\rightarrow$ Conv3x3   $\rightarrow$ AvgPool, skip $\rightarrow$ Conv3x3 $\rightarrow$ \textcolor{red}{skip} \\ \cline{2-4} 
  &
   \textbf{All sizes} &
   \textbf{Model MultiXbar} &
   Conv3x3, Conv5x5 $\rightarrow$ Conv5x5 $\rightarrow$   AvgPool $\rightarrow$ Conv5x5 $\rightarrow$ AvgPool $\rightarrow$ Conv3x3   $\rightarrow$ AvgPool, skip $\rightarrow$ Conv5x5 $\rightarrow$ Conv3x3  \\
 \hline
\end{tabular}%
}
\vspace{-3mm}
\end{table*}

\begin{wrapfigure}{l}{0.5\textwidth}
\includegraphics[width=0.5\textwidth]{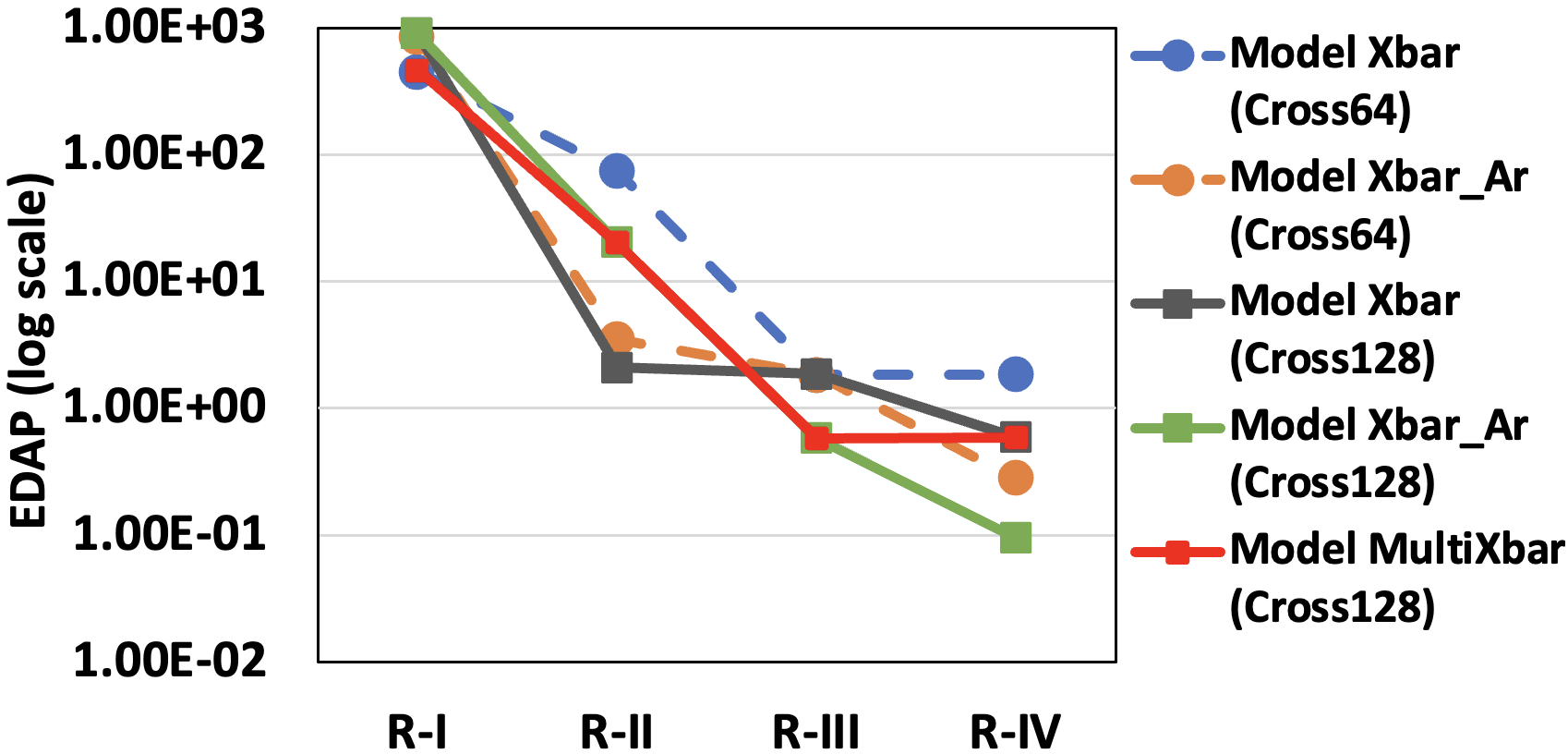}
  
\caption{Plot of EDAPs on crossbars for different residual blocks in Subnets using CIFAR10 dataset.}
\label{hw_res}
\vspace{-4mm}
\end{wrapfigure}

There are a total of nine Op-1 and Op-2 operations in the Supernet (see Fig. \ref{supernet}). Let us collectively refer to them as \textit{Select} operations. In this section, we sequentially arrange the \textit{Select} operations of the different Subnets determined by our \textit{XploreNAS} approach and analyse the architectures (see Table \ref{tab:architectures}). It is evident that a DNN layer corresponding to a 3x3 weight-kernel size would require lower number of crossbars to be mapped than 5x5 weight-kernel size. Further, the latter convolutional layers in the searched Subnets would entail a larger number of crossbars for mapping owing to increase in the number of input/output channels. Based on the results in Table \ref{tab:architectures}, we carry our analysis as follows:

\textbf{Comparison between Model Xbar and Model Xbar\_Ar:} For a given crossbar size, when we compare the architectures of Model Xbar and Model Xbar\_Ar, we find that Model Xbar\_Ar chooses an architecture that requires overall a lower number of crossbars to be mapped. This includes preferring convolutions with lower weight-kernel sizes in the latter residual blocks or replacing convolutional layers with AvgPool or skip-connections. For example, in case of CIFAR10 results for crossbar sizes ranging from 32$\times$32 to 128$\times$128, we find that the last \textit{Select} operation is a skip-connection for Model Xbar\_Ar instead of Conv3x3 for Model Xbar (highlighted in red in Table \ref{tab:architectures}). The trends are similar even for CIFAR100 dataset (data not shown for brevity). Reduction in overall crossbar count brings in higher hardware-efficiency (lower EDAP) to the Model Xbar\_Ar as we can see in Section \ref{sec:el_result}. For the CIFAR10 dataset, we plot blockwise EDAP for the four residual blocks in our Subnets (see Fig. \ref{hw_res}) and find that the major reduction in EDAP for Model Xbar\_Ar arises from the \textit{Select} operations in the latter residual blocks (R-II \& R-IV for 64$\times$64 crossbars and R-III \& R-IV for 128$\times$128 crossbars). Additionally, since \textit{XploreNAS} inherently performs a performance vs. EDAP trade-off when searching for a MultiXbar Subnet, we find that MultiXbar on 128$\times$128 crossbars undergoes reduction in EDAP primarily for the R-I \& R-III residual blocks compared to its Model Xbar counterpart to achieve $\sim1.15\times$ lower overall EDAP.

\textbf{Comparison between different crossbar sizes:} For Model XBar with CIFAR10 dataset, on moving from a crossbar size of 32$\times$32 to 64$\times$64, we find that for the sixth and eighth \textit{Select} operations, Conv5x5 gets replaced with Conv3x3 operation. Similarly, for the second \textit{Select} operation, the ensemble of Conv3x3 \& Conv5x5 operation gets replaced with a single Conv5x5 operation. This has been highlighted in blue in Table \ref{tab:architectures}. Similar trends are also seen in case of Model Xbar\_Ar on moving from the crossbar sizes of 32$\times$32, 64$\times$64 to 128$\times$128. In other words, \textit{XploreNAS} generally opts for convolution operations with smaller weight-kernel sizes to reduce the dimensions of the weight matrices when searching the optimal architecture on larger crossbar sizes. However, on moving to very large crossbar sizes such as 128$\times$128 (having a greater impact of non-idealities), very small weight matrices may result in non-ideality dominance that can adversely affect the performance (robustness) of the DNN model \cite{bhattacharjee2022examining}. Thus, we find that \textit{XploreNAS} chooses a mix of small and large kernel operations to balance the hardware cost alongside performance on 128$\times$128 crossbars (highlighted in magenta in Table \ref{tab:architectures}). Interestingly, \textit{XploreNAS} takes the best of both worlds for Model MultiXbar and yields an optimal mix of smaller and larger kernels that corroborates to its balanced EDAP vs. performance trade-off. 
 \vspace{-3mm}

\begin{figure}[]
    \centering
    
    \includegraphics[width=.8\linewidth]{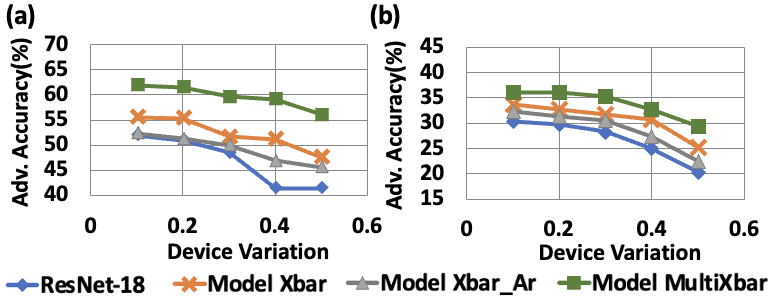}%
    \caption{Plot of variation in PGD-10 adversarial accuracies of Model Xbar, Model Xbar\_Ar, Model MultiXbar and ResNet-18 baseline with NVM device noise on 64$\times$64 crossbars using- (a) CIFAR10 dataset, (b) CIFAR100 dataset.}
    \label{nf_noise_var}
    
\end{figure}

\subsection{Results on varying NVM device noise in crossbars during inference}

So far in all our experiments (for \textit{XploreNAS} and baselines), we have assumed that the NVM devices in the crossbars have synaptic variations with $\sigma/\mu=0.35$, both during training and inference as specified in Section \ref{sec:expt}.
In this section, we infer the models with different levels of crossbar noise by altering the synaptic device variations. Fig. \ref{nf_noise_var}(a-b) show the plots between PGD-10 adversarial accuracy and device variations for CIFAR10 \& CIFAR100 datasets respectively. The device variations are varied from 0.1 to 0.5 for DNNs mapped on 64$\times$64 crossbars during inference.

We find that our \textit{XploreNAS}-derived models to be highly resilient across this range of device noise without re-training; specifically the Model MultiXbar architecture maintains adversarial accuracy in the range of $\sim48-55\%$ for CIFAR10 dataset. In contrast, the standard ResNet-18 baseline model is relatively less robust and suffers higher accuracy losses for device variations over 30\%. Its adversarial accuracy lies in the range of $\sim41-52\%$ for CIFAR10 dataset. Similar trends can also be seen for the CIFAR100 dataset, with our \textit{XploreNAS}-derived Model MultiXbar maintaining its adversarial accuracy in the range of $\sim29-36\%$, while the standard ResNet-18 baseline achieves lower adversarial robustness in the range $\sim20-30\%$.

\section{Comparison with Related Works}
\label{sec:comparison}
\begin{table}[t]

\centering
\caption{Table showing comparison with related works. All quantitative values denote percentage changes with respect to the respective baselines used in these works. Here, '$\times$' denotes non-applicability and $--$ denotes metrics that are not reported in the respective works.}
\label{tab:comparison}
\resizebox{.8\columnwidth}{!}{%
\begin{tabular}{|c|c|c|c|c|c|}
\hline
\textbf{Works} &
  \textbf{Topology} &
  \textbf{\begin{tabular}[c]{@{}c@{}}Xbar \\ Architecture\end{tabular}} &
  \textbf{Non-idealities} &
  \textbf{\begin{tabular}[c]{@{}c@{}}PGD \\ adversarial \\ robustness\end{tabular}} &
  \textbf{\begin{tabular}[c]{@{}c@{}}EDAP \\ improvement \end{tabular}} \\ \hline
\begin{tabular}[c]{@{}c@{}}NEAT, \textit{SwitchX} and others \\ \cite{bhattacharjee2021neat, bhattacharjee2020switchx, roy2020robustness, bhattacharjee2021efficiency}\end{tabular}  & Fixed & Homogeneous   & Present & $+10-11\%$ &
  $+8-22\%$   \\ \hline
\begin{tabular}[c]{@{}c@{}}Guo et al. \\ \cite{guo2020meets}\end{tabular}                & NAS   & $\times$          & $\times$    & $+9-12\%$    & $--$ \\ \hline
\begin{tabular}[c]{@{}c@{}}DANCE for systolic arrays \\ \cite{choi2021dance}\end{tabular} & NAS   & $\times$          & $\times$   & $--$ & $+84-92\%$    \\ \hline
\begin{tabular}[c]{@{}c@{}}NAX for crossbars \\ \cite{negi2021nax}\end{tabular}                       & NAS   & Heterogeneous & Present & $--$ & $+6-46\%$    \\ \hline
\textit{XploreNAS} (Ours)                                                                              & NAS   & Homogeneous   & Present & $+8-16\%$ &
  $+50-60\%$    \\ \hline
\end{tabular}%
}

\end{table}

We provide qualitative \& quantitative comparison between \textit{XploreNAS} and previous works in Table \ref{tab:comparison}. Non-NAS works such as \cite{bhattacharjee2021neat, bhattacharjee2020switchx, roy2020robustness, bhattacharjee2021efficiency} have used algorithm-hardware co-design to achieve adversarial robustness with competitive hardware-efficiencies. Other NAS-based co-design works such as \cite{choi2021dance, negi2021nax} have shown significant EDAP reductions during inference on their respective hardware platforms. But, none of these have focused on optimizing neural architectures for adversarial robustness. Contrarily, \cite{guo2020meets} is a purely hardware-agnostic approach to boost adversarial robustness of DNNs on software and hence, does not deal with hardware metrics such as EDAP. As pointed out in Section \ref{sec:expt}, SW-NAS models derived via \cite{guo2020meets} are highly vulnerable on crossbar platforms and lose their adversarial robustness owing to the non-idealities. \textit{XploreNAS} explores the best of both the worlds and achieves a balanced trade-off between adversarial robustness \& EDAP. 

\section{Conclusion}
\label{sec:conclusion}

In the quest for achieving adversarial security for DNNs deployed on memristive crossbar accelerators, we propose an algorithm-hardware co-optimization approach called \textit{XploreNAS}. It uses one-shot NAS for searching adversarially robust DNNs for non-ideal crossbar platforms. Our experiments show \textit{XploreNAS}-derived Subnets can achieve a balanced trade-off between adversarial robustness and hardware-efficiency in terms of lower EDAP. Furthermore, we also find the Model MultiXbar architectures derived by \textit{XploreNAS} to have adversarial resilience over a range of crossbar sizes and device-level variations without the requirement for re-training. Our \textit{XploreNAS} work can further motivate future studies towards better methods integrated with the Subnet fine-tuning stage for adversarially training or adapting the sampled Subnets to strong adversarial attacks on memristive crossbars.

\begin{acks}
This work was supported in part by C-BRIC, a JUMP center sponsored by DARPA and SRC, CoCoSys, a JUMP2.0 center sponsored by DARPA and SRC, Google Research Scholar Award, the NSF CAREER Award, TII (Abu Dhabi), the DARPA AI Exploration (AIE) program, and the DoE MMICC center SEA-CROGS (Award \#DE-SC0023198).

\end{acks}

\bibliographystyle{ACM-Reference-Format}
\bibliography{sample-base}


\end{document}